\documentclass[10pt,twocolumn,letterpaper]{article}
\usepackage{multirow,colortbl,float}

\usepackage[pagenumbers]{cvpr} 

\usepackage[dvipsnames]{xcolor}

\definecolor{cvprblue}{rgb}{0.21,0.49,0.74}
\usepackage[pagebackref,breaklinks,colorlinks,citecolor=cvprblue]{hyperref}

\title{Mixture of Low-rank Experts for Transferable AI-Generated Image Detection}

\author{
Zihan Liu \qquad Hanyi Wang \qquad Yaoyu Kang \qquad Shilin Wang \\\;\\
Shanghai Jiao Tong University \\
{\tt\small \{lzh123,why\_820,kangyaoyu,wsl\}@sjtu.edu.cn}
}

\begin{document}
\maketitle
\begin{abstract}
\label{sec:abstract}

Generative models have shown a giant leap in synthesizing photo-realistic images with minimal expertise, sparking concerns about the authenticity of online information. This study aims to develop a universal AI-generated image detector capable of identifying images from diverse sources. Existing methods struggle to generalize across unseen generative models when provided with limited sample sources. Inspired by the zero-shot transferability of pre-trained vision-language models, we seek to harness the nontrivial visual-world knowledge and descriptive proficiency of CLIP-ViT to generalize over unknown domains. This paper presents a novel parameter-efficient fine-tuning approach, mixture of low-rank experts, to fully exploit CLIP-ViT's potential while preserving knowledge and expanding capacity for transferable detection. We adapt only the MLP layers of deeper ViT blocks via an integration of shared and separate LoRAs within an MoE-based structure. Extensive experiments on public benchmarks show that our method achieves superiority over state-of-the-art approaches in cross-generator generalization and robustness to perturbations. Remarkably, our best-performing ViT-L/14 variant requires training only 0.08\% of its parameters to surpass the leading baseline by +3.64\% mAP and +12.72\% avg.Acc across unseen diffusion and autoregressive models. This even outperforms the baseline with just 0.28\% of the training data. Our code and pre-trained models will be available at \url{https://github.com/zhliuworks/CLIPMoLE}.
\end{abstract}    
\section{Introduction}
\label{sec:introduction}

Recent advances in diffusion models and natural language processing facilitate the growth of powerful image synthesis tools (\textit{e.g.} Stable Diffusion~\cite{22sd}, Midjourney~\cite{midjourney}). Their public releases empower users to create photo-realistic images with unprecedented control. However, these tools require minimal expertise and considerably lower the barrier for image manipulation. This raises severe concerns about the credibility of images shared over online platforms, including deepfakes, impersonation, and copyright infringement. These platforms are overwhelmed with untrustworthy AI-generated content, necessitating effective countermeasures to detect potential synthetic images.

\begin{figure}[t]
  \centering
   \includegraphics[width=\linewidth]{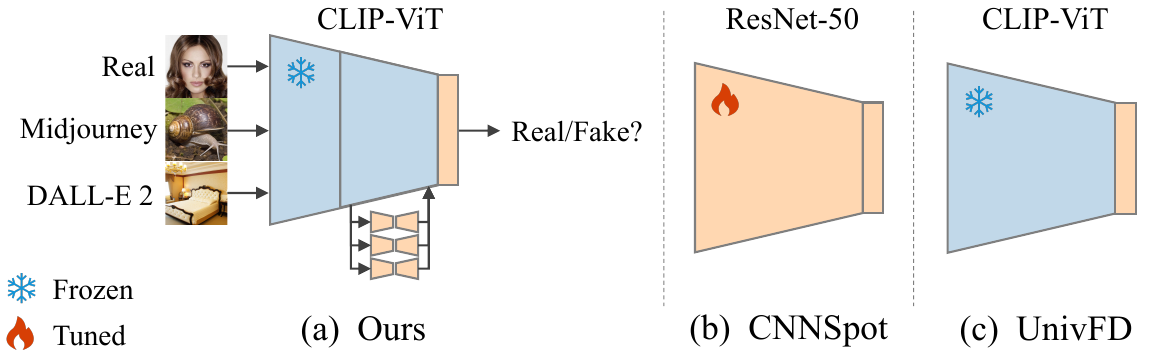}
   \caption{\textbf{Fine-tuning methods comparison.} Compared to (b) full fine-tuning~\cite{20wang} and (c) linear probing~\cite{23universal}, our method allows for more effective and efficient adaption of CLIP-ViT for this task.}
   \label{fig:intro}
\end{figure}

Existing generated image detectors mainly focus on images produced by GANs. Many of these methods capture the underlying traces left by GANs in spatial~\cite{19fingerprint,19attribute,19ffpp} and frequency~\cite{19zhang,20frank,21closer,22twice} domains. However, deep networks often overfit to the low-level artifacts in the training set, hindering generalization over unseen generative models. Hence, developing a universal AI-generated image detector has attracted considerable interests~\cite{20wang,20patch}, particularly for the latest diffusion models~\cite{23universal,23dire}. Wang \textit{et al.}~\cite{20wang} highlight the importance of sufficient training data and appropriate augmentations for transferability. Ojha \textit{et al.}~\cite{23universal} show promising generalization by using the pre-trained features of vision-language models without explicit training. Following their setup, we aim to design a transferable detection system to identify any source of fake images, given access to images from a single generative model.

This study assumes intrinsic discrepancies between real and fake images particularly at low-level forensic details, such as textural~\cite{20texture} and spectral traces~\cite{22byreal}, with different generators exhibiting properties of their unique natures. However, we argue that intentionally introducing these biases could cause the model to easily overfit, thereby overlooking any inherent features of natural images and restricting its ability to generalize. Motivated by the significant benefits of off-the-shelf vision-language models in zero-shot scenarios, we aim to embrace the open-world visual knowledge and descriptive proficiency of CLIP-ViT~\cite{21clip}, a contrastively pre-trained visual encoder exposed to a web-scale collection of image-text pairs. In order to effectively tailor CLIP-ViT for our task while ensuring modest degradation of its pre-trained features, we propose a parameter-efficient fine-tuning method, mixture of low-rank experts, as in Figure~\ref{fig:intro}. This approach is designed to fully exploit the potential of CLIP-ViT, enabling robust and generalizable detection across images of unknown domains.

Specifically, we freeze the backbone parameters to enable low-rank adapters (LoRA)~\cite{22lora} to access the visual knowledge preserved in CLIP-ViT. Shallow blocks capture general low-level properties without additional learning, which are fed to deeper blocks for more fine-grained forensic analysis. The MLP layers of deeper blocks are adapted using a combination of shared and separate LoRAs within an mixture-of-experts (MoE) based framework. Shared adapters learn common feature prototypes from the entire dataset, whereas separate adapters are sparsely activated to specialize in varied generative patterns of synthetic images. The MoE structure expands the capacity of adapters by increasing learnable parameters to model multifaceted natures in both natural and generated images. This allows multiple experts to jointly approximate the weight updates without over-reliance on any specific adapter, promoting generalization similar to the idea of dropout~\cite{14dropout,23moedropout}.

Extensive experiments on UnivFD~\cite{23universal} and GenImage~\cite{23genimage} benchmarks show the superiority of our method over state-of-the-art approaches in terms of generalization across unseen generators (Sec.~\ref{sec:generalize}) and robustness to post-processing operations (Sec.~\ref{sec:robust}). For instance, when training on 720k ProGAN/LSUN dataset~\cite{20wang} and evaluating on unseen diffusion and autoregressive models, we outperform the leading baseline UnivFD~\cite{23universal} by +3.64\% mAP and +12.72\% avg.Acc with the same backbone (\textit{i.e.} CLIP ViT-L/14~\cite{21clip}). Notably, we surpass it even with only 0.28\% of the training data. Besides, the training process is computationally efficient and resource-friendly as only 0.08\% of the parameters are optimized. We also empirically examine the ingredients that contribute to its performance through ablations. For example, is adaptation requisite for every ViT block? (Sec.~\ref{sec:ablation_block}) What rationale underlies the specific design choices made for our adapter modules? (Sec.~\ref{sec:ablation_adapter}) Do LoRA ranks notably impact performance? (Sec.~\ref{sec:ablation_rank})

Our main contributions are: (1) We propose a parameter-efficient fine-tuning method, mixture of low-rank experts, to tailor CLIP-ViT for transferable AI-generated image detection. Our key takeaway is to adapt the MLP layers of the deeper blocks with mixture of shared and separate LoRA experts. (2) Extensive experiments show that our method achieves state-of-the-art performance in cross-generator generalization and robustness to perturbations. We provide a thorough analysis of the crucial ingredients that contribute to the strong transferability of our method. 

\section{Related work}
\label{sec:relatedwork}

\noindent\textbf{Image Generation.}
Image generation relies on deep generative models, aiming to characterize the intricate data distribution of natural images. Recent advances include generative adversarial networks (GAN)~\cite{14gan}, variational autoencoders~\cite{14vae}, autoregressive models~\cite{20imagegpt}, normalizing flows~\cite{17flow}, and diffusion models~\cite{20ddpm}. GANs, notably the ProGAN/StyleGAN family~\cite{18progan,19stylegan}, excel at efficiently generating high-resolution images with impressive perceptual quality. Recently, diffusion models have greatly enhanced image generation quality in terms of fidelity, diversity, and controllability~\cite{21adm,22pndm,22cfg}. Cutting-edge releases such as Stable Diffusion~\cite{22sd}, DALL-E 2~\cite{22dalle2}, and Imagen~\cite{22imagen} showcase the capacity to produce ultra-realistic images from textual prompts, gradually blurring the line between natural and AI-generated images.\\

\noindent\textbf{Generated Image Detection.}
Early image forensics detect hand-crafted cues from traditional tools, such as artifacts~\cite{13shadow,17jpeg} and operations~\cite{18splicedetect,19pswarp}. Recent methods seek to leverge deep networks to identify traces left by generative models in spatial or frequency domains, focusing on images generated by GANs. Spatial-domain approaches include co-occurrence matrices~\cite{19cooccurrence}, unique fingerprints~\cite{19attribute}, Gram matrices~\cite{20texture}, local patches~\cite{20patch}, and \textit{etc}. Frequency analysis of GAN-generated images is also a prominent line of research, exploring spectral artifacts~\cite{19zhang,20frank} and power distribution discrepancies~\cite{20durall,20dzanic,21closer}. Zhang \textit{et al.}~\cite{19zhang} propose a generator to simulate GAN sampling artifacts for generalizable spectrum detection. Frank \textit{et al.}~\cite{20frank} suggest that spectral artifacts in GAN-generated images stem from standard upsampling operations used in GANs.

However, generalization over newer breeds of generative models remains challenging due to learned asymmetric decision boundary~\cite{23universal} and continuous architectural changes~\cite{23online}. Detectors are prone to overfitting to subtle artifacts specific to the training set, leading to suboptimal performance on unseen generators. Wang \textit{et al.}~\cite{20wang} present an effective baseline by training a ResNet-50~\cite{16resnet} classifier on 720k ProGAN~\cite{18progan} and LSUN~\cite{15lsun} images with Gaussian blur and JPEG compression, which exhibits impressive generalization in detecting other GANs such as StyleGAN~\cite{19stylegan} and CycleGAN~\cite{17cyclegan}. More recently, many works are focusing on the detection of diffusion-generated images~\cite{23universal,23dire,23online,23genimage,23intriguing}. Ojha \textit{et al.}~\cite{23universal} implement nearest neighbor and linear probing classifiers on the pre-trained visual features of CLIP~\cite{21clip} to show strong generalization. Their promising results inspire us to delve deeper into the potential of CLIP-ViT for this task.\\

\noindent\textbf{Low-Rank Adaption.}
Low-Rank Adaption (LoRA)~\cite{22lora} is a prevalent method for efficiently fine-tuning pre-trained models for downstream tasks. When data is limited, fine-tuning the entire model can lead to catastrophic forgetting of pre-trained knowledge. LoRA hypothesizes that the updates to model weights reside on a low intrinsic dimension during fine-tuning, and injects trainable low-rank matrices to approximate the weight updates. LoRA only trains this small set of additional parameters, preserving the versatility of the original model while effectively adapting it to specific tasks. It also improves on computational efficiency and resource usage.\\

\noindent\textbf{Mixture of Experts.}
Mixture-of-Experts (MoE)~\cite{17outrageous,21gshard,22switchtransformer} significantly expands model capacity without a proportional increase in computational costs. Compared to the vanilla Transformer, MoE replaces the MLP layer in each block with an MoE layer. Each MoE layer contains multiple independent sub-networks as experts, coupled with a trainable gating function as a router to determine a sparse combination of these experts to use for each example.

\section{Method}
\label{sec:method}

\subsection{Motivations}
Our goal is to develop a universal AI-generated image detector endowed with strong transferability across unseen generators. Previous studies~\cite{23universal,23lgrad} have observed a generalization bottleneck resulting from detectors focusing only on the presence/absence of low-level artifacts to discriminate real and fake images, overlooking the properties inherent in natural images.

\begin{figure}[t]
  \centering
   \includegraphics[width=\linewidth]{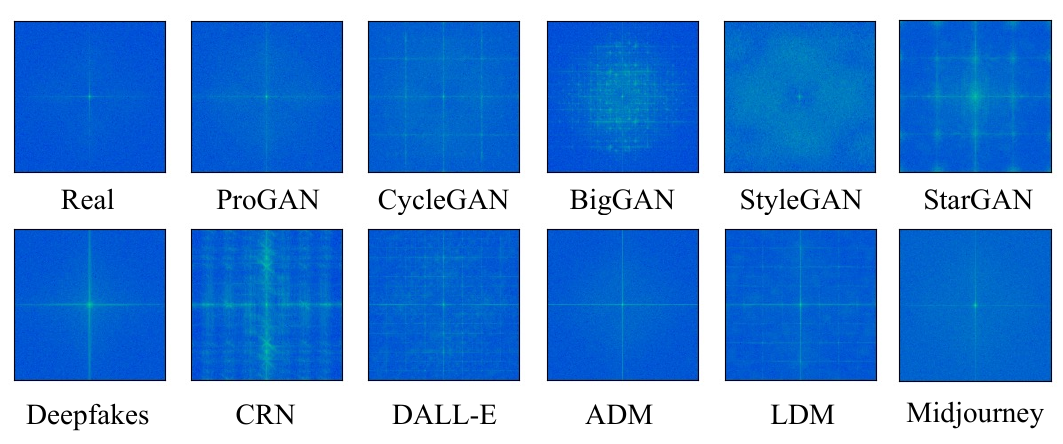}
   
   \caption{\textbf{Average FFT spectra of the high-pass filtered images.} The first one represents real images from ImageNet~\cite{09imagenet}. The last 11 spectra correspond to distinct classes of fake images.}
 
   \label{fig:freq_spec}
\end{figure}

For low-level analysis, we study the frequency distribution of several generators. Following~\cite{19zhang,20wang}, we visualize the average Fast Fourier Transform (FFT) spectra of the high-frequency components of different image distributions. Figure~\ref{fig:freq_spec} depicts the spectra for real images (from ImageNet~\cite{09imagenet}) and 11 categories of fake images (from UnivFD~\cite{23universal} and GenImage~\cite{23genimage} datasets). We observe both similarities and disparities among the generators in terms of periodic patterns and low-frequency power. For instance, Deepfakes~\cite{19ffpp} and ADM~\cite{21adm} exhibit more pronounced low-frequency deviations from real images (at the center of the spectra), and LDM~\cite{22sd} and DALL-E~\cite{21dalle} display analogous grid-like patterns.

However, switching to a backbone of higher capacity fails to effectively address the generalization issue, since it is more prone to overfitting to such artifacts appeared in the training set~\cite{23universal}. In this case, we seek to leverage pre-trained vision-language models to push the generalization ceiling of the detector further.\\

\noindent\textbf{Visual-world knowledge and descriptive power facilitate generalization over unknown domains.}
Large vision-language models store valuable visual knowledge that can considerably enhance generalization over out-of-distribution scenarios, particularly with limited sample sources. These models have been trained on web-scale collections of image-text pairs, offering semantic-rich representations of open-world visual concepts crucial for handling unknown generators. Moreover, the image-text alignment pre-training task equips them with powerful descriptive capabilities, enabling better perception of low-level local details essential for discriminating real from fake images~\cite{20patch}. Therefore, we choose CLIP-ViT~\cite{21clip} as backbone, a contrastively pre-trained visual encoder exposed to 400M image-text pairs, which shows nontrivial zero-shot transfer abilities on downstream tasks. Relevant methods \cite{22defake,23universal} simply freeze the entire CLIP and train a classifier using its features. However, the general encoder lacks task-specific adaptation and struggles with limited learnable parameters, significantly constraining its capacity and hindering further improvements in generalization.\\

\begin{figure*}[t]
  \centering
   \includegraphics[width=.95\linewidth]{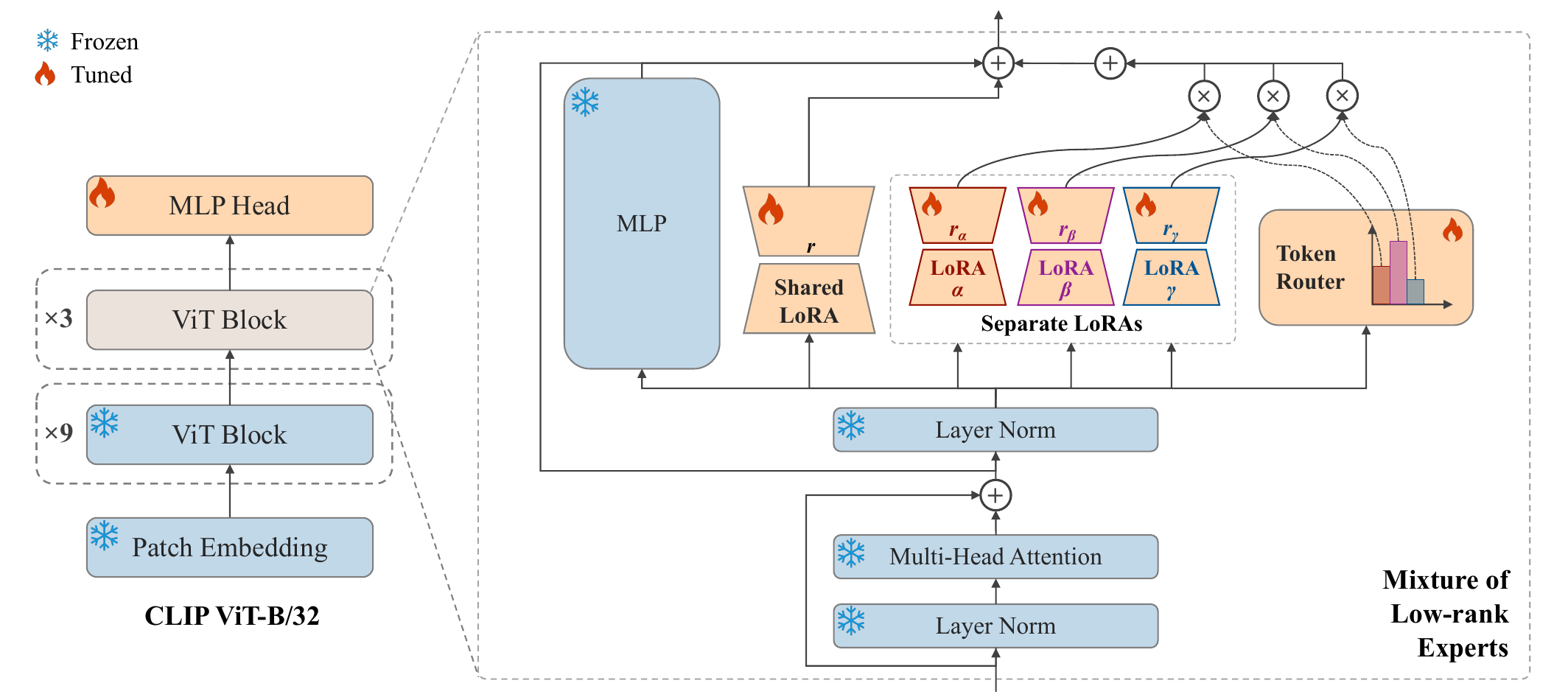}

   \caption{\textbf{Overview of our proposed mixture of low-rank experts for AI-generated image detection.} For the last three blocks of CLIP ViT-B/32, we introduce an integration of shared and separate LoRAs of different ranks as adaptable low-rank experts. The router is responsible for assigning each token to one separate LoRA expert in the MLP layer of each block. During fine-tuning, only the LoRA experts, the routers, and the MLP head are optimized.}
   
   \label{fig:overview}
\end{figure*}

\noindent\textbf{How to fully exploit the aptitude of CLIP-ViT for transferable detection?}
Our key motivation is to customize more CLIP parameters for our task while mitigating the catastrophic forgetting of its pre-trained knowledge. A straightforward way is to unfreeze all or part of the parameters for fine-tuning, but this could lead to performance degradation for two factors: On one hand, with an extensive set of pre-trained CLIP parameters and limited training data, fine-tuning is prone to overfitting. On the other hand, fine-tuning could distort CLIP's pre-trained features, resulting in decreased performance when faced with significant distribution shift~\cite{22ftood}, as observed in our scenario.

Recent success in parameter-efficient transfer learning motivates us to train task-specific adapter modules while keeping the pre-trained parameters frozen to retain knowledge. We avoid the intentional use of adapters that introduce forgery-aware priors~\cite{23dfadapter,23forgeryawaredeepfake} to prevent any subjective biases from affecting the low-level forensics of CLIP. Therefore, we use the widely accepted LoRA~\cite{22lora} to implement our initial idea: adapting CLIP-ViT by directly fine-tuning the MLP layers within Transformer blocks using LoRA. We prioritize adapting the MLP layers since they operate as key-value memories in Transformer-based language models~\cite{21memory}, thus storing visual knowledge in ViT. Surprisingly, we observe that such tuned model already performs considerably well (Sec.~\ref{sec:ablation_adapter} for details).

However, we argue that CLIP's potential has not been fully unleashed via a simple LoRA, as indicated by varying optimal LoRA ranks across different tested generative models (Sec.~\ref{sec:ablation_rank} for details). In MoE architectures, the gating function selects appropriate experts based on data properties, allowing collaboration in completing the forward process. Inspired by this, we suggest incorporating multiple LoRA experts of different ranks to further expand capacity while retaining original strengths as a general visual encoder. This MoE-based approach enables diverse low-rank experts to identify distinct patterns in both natural and AI-generated images through the routing mechanism. This prevents over-reliance on the LoRA of a specific rank and enhances generalization similar to the idea of dropout~\cite{14dropout,23moedropout}. Meanwhile, it maintains computational efficiency during fine-tuning by avoiding introducing substantial parameters.

\subsection{Mixture of low-rank experts for CLIP-ViT}
We propose a simple and effective parameter-efficient fine-tuning method, mixture of low-rank experts, to boost CLIP-ViT's transferability for AI-generated image detection. Our approach applies to various CLIP variants, with model performance scaling in proportion to the number of parameters. As an illustration, we use CLIP ViT-B/32, the most lightweight version for most experiments. Specifically, we freeze the backbone parameters to allow low-rank experts to access visual knowledge in the base model. Shallow blocks tend to easily overfit to undesirable artifacts from specific generative models in the training set, leading to poor generalization since deeper blocks struggle to learn intricate patterns. Therefore, we avoid optimizing the parameters of shallow blocks to ensure no degradation of CLIP-ViT's low-level perception and preserve more general cues for forensic analysis within deeper blocks.

As depicted in Figure~\ref{fig:overview}, for each MLP layer of the deeper ViT blocks (\textit{i.e.} the last three blocks of ViT-B/32), we define a shared LoRA and multiple separate LoRAs of different ranks as adaptable low-rank experts, connected by a router. Shared LoRA acquires mutual feature prototypes from the entire dataset, while the sparse mixture of separate LoRAs promotes specialization in diverse generative patterns of AI-generated images. For instance, shared and separate LoRAs prioritize the similarities and disparities among the spectra of the generators, as illustrated in Figure~\ref{fig:freq_spec}.

These experts are trainable and serve as bypasses for the MLP layers. The router is a learnable linear layer that routes each patch token to one separate LoRA with the highest gated probability. It applies load balancing loss on each mixture layer to prevent all inputs from being dispatched to the same expert. Finally, we add a single linear layer with sigmoid activation as the MLP head. We train only these newly inserted modules for real-vs-fake classification.

Formally, for each adapted block, let $\boldsymbol{x}\in\mathbb{R}^d$ denotes the input to the MLP layer with dimension $d$. During training, the matrix operation within the MLP layer can be represented as: $
\boldsymbol{h}=\boldsymbol{W_0}\boldsymbol{x}+\boldsymbol{\Delta W}\boldsymbol{x}
$, where $\boldsymbol{W_0}\in\mathbb{R}^{d\times d}$ is the weight matrix and $\boldsymbol{\Delta W}\in\mathbb{R}^{d\times d}$ is the weight update. $\boldsymbol{h}\in\mathbb{R}^d$ is the output of this block. To estimate $\boldsymbol{\Delta W}$, we devise a linear combination of shared and $N$ separate low-rank experts. The forward process can be expressed as:
\[
\boldsymbol{\Delta W}\boldsymbol{x}=\frac{\alpha}{r}\boldsymbol{B}\boldsymbol{A}\boldsymbol{x}+\sum_{i=1}^NG_i(\boldsymbol{x})\frac{\alpha_i}{r_i}\boldsymbol{B_i}\boldsymbol{A_i}\boldsymbol{x}
\]
where $G_i(\boldsymbol{x})=\text{Softmax}(\boldsymbol{W_gx})_i$ is the router function to determine the single activated separate LoRA. $\boldsymbol{W_g}\in\mathbb{R}^{N\times d}$ is the trainable matrix of the router. The shared LoRA comprises a pair of low-rank matrices, $\boldsymbol{B}\in\mathbb{R}^{d\times r}$ and $\boldsymbol{A}\in\mathbb{R}^{r\times d}$, where $r$ ($\ll d$) and $\alpha$ are the rank and scaling constant, respectively. Similarly, each separate LoRA consists of $\boldsymbol{B_i}\in\mathbb{R}^{d\times r_i}$ and $\boldsymbol{A_i}\in\mathbb{R}^{r_i\times d}$, where $r_i$ ($\ll d$) and $\alpha_i$ are the corresponding rank and scaling constant.\\

\noindent\textbf{Loss function.}
Using only binary cross-entropy loss $\mathcal{L}_\text{bce}$ can lead to severe load imbalance, where $\boldsymbol{W_g}$ collapses such that almost all the inputs are assigned to the same expert. To mitigate this, we introduce an auxiliary loss~\cite{22switchtransformer} to encourage uniform dispatching across experts. For the $i$-th adapted block, given the MLP input $\boldsymbol{x}$, $N$ separate low-rank experts, and a batch $\mathcal{B}$ with $T$ tokens, the load balancing loss is computed as:
\begin{align*}
\mathcal{L}_\text{lb}^i &= N\sum_{j=1}^Nf_j^iP_j^i \\
f_j^i &= \frac{1}{T}\sum_{\boldsymbol{x}\in\mathcal{B}}\mathbf{1}\{\text{argmax}\;G_j^i(\boldsymbol{x})=j\} \\
P_j^i &= \frac{1}{T}\sum_{\boldsymbol{x}\in\mathcal{B}}G_j^i(\boldsymbol{x})
\end{align*}
where $f_j^i$ is the fraction of tokens routed to the $j$-th expert, and $P_j^i$ is the fraction of the gated probability of the $j$-th expert. $\mathcal{L}_\text{lb}^i$ is optimized under a uniform distribution. Finally, the total model loss $\mathcal{L}$ is defined as:
\[
\mathcal{L}=\mathcal{L}_\text{bce}+\lambda\sum_{i\in\mathcal{D}}\mathcal{L}_\text{lb}^i
\]
where $\mathcal{D}$ is the set of blocks to adapt, and $\lambda$ is a hyper-parameter to regulate the effect of load balancing.
\section{Experiments}
\label{sec:experiments}

\begin{table*}[htbp]
\centering
\resizebox{\linewidth}{!}{
\begin{tabular}{lcccccccccccccccccccc}
\toprule
   &
  \multicolumn{6}{c}{Generative adversarial networks} &
   &
  \multicolumn{2}{c}{Low level vision} &
  \multicolumn{2}{c}{Perceptual loss} &
  \multicolumn{7}{c}{Diffusion models} & &
  Total \\ \cmidrule(lr){2-7} \cmidrule(lr){9-10} \cmidrule(lr){11-12} \cmidrule(lr){13-19} \cmidrule(l){21-21}
  
   &
   &
   &
   &
   &
   &
   &
   &
   &
   &
   &
   &
   &
  \multicolumn{3}{c}{LDM} &
  \multicolumn{3}{c}{GLIDE} &
   &
   \\ \cmidrule(lr){14-16} \cmidrule(lr){17-19}
  \multirow{-3}{*}{Method} &
  \multirow{-2}{*}{\begin{tabular}[c]{@{}c@{}}Pro-\\ GAN\end{tabular}} &
  \multirow{-2}{*}{\begin{tabular}[c]{@{}c@{}}Cycle-\\ GAN\end{tabular}} &
  \multirow{-2}{*}{\begin{tabular}[c]{@{}c@{}}Big-\\ GAN\end{tabular}} &
  \multirow{-2}{*}{\begin{tabular}[c]{@{}c@{}}Style-\\ GAN\end{tabular}} &
  \multirow{-2}{*}{\begin{tabular}[c]{@{}c@{}}Gau-\\ GAN\end{tabular}} &
  \multirow{-2}{*}{\begin{tabular}[c]{@{}c@{}}Star-\\ GAN\end{tabular}} &
  \multirow{-3}{*}{\begin{tabular}[c]{@{}c@{}}Deep-\\ fakes\end{tabular}} &
  \multirow{-2}{*}{SITD} &
  \multirow{-2}{*}{SAN} &
  \multirow{-2}{*}{CRN} &
  \multirow{-2}{*}{IMLE} &
  \multirow{-2}{*}{Guided} &
  200s &
  \begin{tabular}[c]{@{}c@{}}200s\\ w/CFG\end{tabular} &
  100s &
  \begin{tabular}[c]{@{}c@{}}100-\\ 27\end{tabular} &
  \begin{tabular}[c]{@{}c@{}}50-\\ 27\end{tabular} &
  \begin{tabular}[c]{@{}c@{}}100-\\ 10\end{tabular} &
  \multirow{-3}{*}{DALL-E} &
  \multirow{-2}{*}{mAP} \\ \midrule

CNNSpot~\cite{20wang} &
  \textbf{100.0} &
  93.47 &
  84.50 &
  99.54 &
  89.49 &
  98.15 &
  89.02 &
  73.75 &
  59.47 &
  \textbf{98.24} &
  98.40 &
  73.72 &
  70.62 &
  71.00 &
  70.54 &
  80.65 &
  84.91 &
  82.07 &
  70.59 &
  83.58 \\ \midrule
  
PatchForensics~\cite{20patch} &
  80.88 &
  72.84 &
  71.66 &
  85.75 &
  65.99 &
  69.25 &
  76.55 &
  76.19 &
  76.34 &
  74.52 &
  68.52 &
  75.03 &
  87.10 &
  86.72 &
  86.40 &
  85.37 &
  83.73 &
  78.38 &
  75.67 &
  77.73 \\ \midrule
  
CoOccurrence~\cite{19cooccurrence} &
  \underline{99.74} &
  80.95 &
  50.61 &
  98.63 &
  53.11 &
  67.99 &
  59.14 &
  68.98 &
  60.42 &
  73.06 &
  87.21 &
  70.20 &
  91.21 &
  89.02 &
  92.39 &
  89.32 &
  88.35 &
  82.79 &
  80.96 &
  78.11 \\ \midrule
  
FreqSpec~\cite{19zhang} &
  55.39 &
  \textbf{100.0} &
  75.08 &
  55.11 &
  66.08 &
  \textbf{100.0} &
  45.18 &
  47.46 &
  57.12 &
  53.61 &
  50.98 &
  57.72 &
  77.72 &
  77.25 &
  76.47 &
  68.58 &
  64.58 &
  61.92 &
  67.77 &
  66.21 \\ \midrule
  
DIRE*~\cite{23dire} &
\textbf{100.0} &
76.73 &
72.80 &
97.06 &
68.44 &
\textbf{100.0} &
\textbf{98.55} &
54.51 &
65.62 &
\underline{97.10} &
93.74 &
\underline{94.29} &
95.17 &
\underline{95.43} &
95.77 &
\underline{96.18} &
\underline{97.30} &
\underline{97.53} &
68.73 &
87.63
   \\ \midrule
   
UnivFD~\cite{23universal} &
  \textbf{100.0} &
  99.46 &
  99.59 &
  97.24 &
  \underline{99.98} &
  99.60 &
  82.45 &
  61.32 &
  79.02 &
  96.72 &
  99.00 &
  87.77 &
  99.14 &
  92.15 &
  99.17 &
  94.74 &
  95.34 &
  94.57 &
  97.15 &
  93.38 \\ \midrule

\rowcolor[HTML]{EFEFEF} 
Ours (ViT-B/32) &
\textbf{100.0} &
99.44 &
98.12 &
98.97 &
97.75 &
99.85 &
79.55 &
77.25 &
72.67 &
76.39 &
95.32 &
93.28 &
98.99 &
94.93 &
99.14 &
91.90 &
93.23 &
92.06 &
96.64 &
92.39
   \\  
   
\rowcolor[HTML]{EFEFEF} 
Ours (ViT-B/16) &
\textbf{100.0} &
99.56 &
\underline{99.66} &
\underline{99.78} &
99.94 &
\underline{99.86} &
85.03 &
\underline{87.95} &
\underline{80.99} &
93.41 &
\underline{99.27} &
89.93 &
\underline{99.46} &
95.06 &
\underline{99.55} &
94.75 &
95.66 &
95.00 &
\underline{97.93} &
\underline{95.41}
   \\ 
   
\rowcolor[HTML]{EFEFEF} 
Ours (ViT-L/14) &
\textbf{100.0} &
\underline{99.97} &
\textbf{99.93} &
\textbf{99.81} &
\textbf{100.0} &
99.81 &
\underline{91.45} &
\textbf{90.93} &
\textbf{84.95} &
96.09 &
\textbf{99.57} &
\textbf{94.40} &
\textbf{99.87} &
\textbf{97.85} &
\textbf{99.93} &
\textbf{99.12} &
\textbf{99.48} &
\textbf{99.23} &
\textbf{99.22} &
\textbf{97.45}
   \\ \bottomrule
\end{tabular}}

\caption{\textbf{Generalization results on UnivFD dataset~\cite{23universal}.} The methods are trained on ProGAN data from~\cite{20wang} (except FreqSpec~\cite{19zhang} using CycleGAN) and evaluated using the average precision (AP) metric. The best result is highlighted in bold, and the second-best is underlined. * denotes our implementation based on official codes.}
\label{tab:general_univfd_ap}
\end{table*}

\begin{table*}[htbp]
\centering
\resizebox{\linewidth}{!}{
\begin{tabular}{lcccccccccccccccccccc}
\toprule
   &
  \multicolumn{6}{c}{Generative adversarial networks} &
   &
  \multicolumn{2}{c}{Low level vision} &
  \multicolumn{2}{c}{Perceptual loss} &
  \multicolumn{7}{c}{Diffusion models} & &
  Total \\ \cmidrule(lr){2-7} \cmidrule(lr){9-10} \cmidrule(lr){11-12} \cmidrule(lr){13-19} \cmidrule(l){21-21}
  
   &
   &
   &
   &
   &
   &
   &
   &
   &
   &
   &
   &
   &
  \multicolumn{3}{c}{LDM} &
  \multicolumn{3}{c}{GLIDE} &
   &
   \\ \cmidrule(lr){14-16} \cmidrule(lr){17-19}
  \multirow{-3}{*}{Method} &
  \multirow{-2}{*}{\begin{tabular}[c]{@{}c@{}}Pro-\\ GAN\end{tabular}} &
  \multirow{-2}{*}{\begin{tabular}[c]{@{}c@{}}Cycle-\\ GAN\end{tabular}} &
  \multirow{-2}{*}{\begin{tabular}[c]{@{}c@{}}Big-\\ GAN\end{tabular}} &
  \multirow{-2}{*}{\begin{tabular}[c]{@{}c@{}}Style-\\ GAN\end{tabular}} &
  \multirow{-2}{*}{\begin{tabular}[c]{@{}c@{}}Gau-\\ GAN\end{tabular}} &
  \multirow{-2}{*}{\begin{tabular}[c]{@{}c@{}}Star-\\ GAN\end{tabular}} &
  \multirow{-3}{*}{\begin{tabular}[c]{@{}c@{}}Deep-\\ fakes\end{tabular}} &
  \multirow{-2}{*}{SITD} &
  \multirow{-2}{*}{SAN} &
  \multirow{-2}{*}{CRN} &
  \multirow{-2}{*}{IMLE} &
  \multirow{-2}{*}{Guided} &
  200s &
  \begin{tabular}[c]{@{}c@{}}200s\\ w/CFG\end{tabular} &
  100s &
  \begin{tabular}[c]{@{}c@{}}100-\\ 27\end{tabular} &
  \begin{tabular}[c]{@{}c@{}}50-\\ 27\end{tabular} &
  \begin{tabular}[c]{@{}c@{}}100-\\ 10\end{tabular} &
  \multirow{-3}{*}{DALL-E} &
  \multirow{-2}{*}{\begin{tabular}[c]{@{}c@{}}avg.\\ Acc\end{tabular}} \\ \midrule

CNNSpot~\cite{20wang} &
  99.99 &
  85.20 &
  70.20 &
  85.70 &
  78.95 &
  91.70 &
  53.47 &
  66.67 &
  48.69 &
  \textbf{86.31} &
  86.26 &
  60.07 &
  54.03 &
  54.96 &
  54.14 &
  60.78 &
  63.80 &
  65.66 &
  55.58 &
  69.58
   \\ \midrule
  
PatchForensics~\cite{20patch} &
  75.03 &
  68.97 &
  68.47 &
  79.16 &
  64.23 &
  63.94 &
  75.54 &
  75.14 &
  \textbf{75.28} &
  72.33 &
  55.30 &
  67.41 &
  76.50 &
  76.10 &
  75.77 &
  74.81 &
  73.28 &
  68.52 &
  67.91 &
  71.24
   \\ \midrule
  
CoOccurrence~\cite{19cooccurrence} &
  97.70 &
  63.15 &
  53.75 &
  92.50 &
  51.10 &
  54.70 &
  57.10 &
  63.06 &
  55.85 &
  65.65 &
  65.80 &
  60.50 &
  70.70 &
  70.55 &
  71.00 &
  70.25 &
  69.60 &
  69.90 &
  67.55 &
  66.86
   \\ \midrule
  
FreqSpec~\cite{19zhang} &
  49.90 &
  \textbf{99.90} &
  50.50 &
  49.90 &
  50.30 &
  \underline{99.70} &
  50.10 &
  50.00 &
  48.00 &
  50.60 &
  50.10 &
  50.90 &
  50.40 &
  50.40 &
  50.30 &
  51.70 &
  51.40 &
  50.40 &
  50.00 &
  55.45
   \\ \midrule
  
DIRE*~\cite{23dire} &
\textbf{100.0} &
67.73 &
64.78 &
83.08 &
65.30 &
\textbf{100.0} &
\textbf{94.75} &
57.62 &
60.96 &
62.36 &
62.31 &
83.20 &
82.70 &
84.05 &
84.25 &
\underline{87.10} &
\underline{90.80} &
\underline{90.25} &
58.75 &
77.89
   \\ \midrule
   
UnivFD~\cite{23universal} &
  \textbf{100.0} &
  98.50 &
  94.50 &
  82.00 &
  \underline{99.50} &
  97.00 &
  66.60 &
  63.00 &
  57.50 &
  59.50 &
  72.00 &
  70.03 &
  \underline{94.19} &
  73.76 &
  94.36 &
  79.07 &
  79.85 &
  78.14 &
  86.78 &
  81.38
 \\ \midrule

\rowcolor[HTML]{EFEFEF} 
Ours (ViT-B/32) &
\textbf{100.0} &
94.38 &
94.53 &
93.66 &
96.26 &
95.45 &
72.41 &
74.44 &
61.87 &
75.26 &
81.72 &
\underline{85.60} &
90.50 &
77.55 &
91.70 &
77.55 &
80.30 &
78.55 &
85.65 &
84.60 \\  
 
\rowcolor[HTML]{EFEFEF} 
Ours (ViT-B/16) &
\textbf{100.0} &
96.95 &
\underline{98.25} &
\underline{95.00} &
99.10 &
96.75 &
75.40 &
\underline{78.72} &
63.27 &
78.76 &
\underline{91.56} &
80.70 &
93.75 &
\underline{86.30} &
\underline{96.25} &
85.50 &
87.40 &
86.85 &
\underline{91.85} &
\underline{88.55} \\ 
   
\rowcolor[HTML]{EFEFEF} 
Ours (ViT-L/14) &
\textbf{100.0} &
\underline{99.33} &
\textbf{99.67} &
\textbf{99.46} &
\textbf{99.83} &
97.07 &
\underline{77.53} &
\textbf{81.11} &
\underline{65.50} &
\underline{82.32} &
\textbf{96.79} &
\textbf{90.70} &
\textbf{98.30} &
\textbf{95.90} &
\textbf{98.75} &
\textbf{92.40} &
\textbf{93.95} &
\textbf{93.00} &
\textbf{94.90} &
\textbf{92.45} \\ \bottomrule
\end{tabular}}

\caption{\textbf{Generalization results on UnivFD dataset~\cite{23universal}.} Analogous results of Table~\ref{tab:general_univfd_ap}, where the methods are evaluated using the classification accuracy (Acc) over real and fake images from each generative model.}
\label{tab:general_univfd_acc}
\end{table*}

\subsection{Experimental setup}
\noindent\textbf{Implementation details.}
We adapt the last three blocks for three variants of CLIP-ViT (\textit{i.e.} ViT-B/32, ViT-B/16, ViT-L/14). The shared LoRA has a rank of 8, and the separate low-rank experts consist of three LoRAs with different ranks, set as $r_\alpha$ = 4, $r_\beta$ = 8, $r_\gamma$ = 16. Consistent with~\cite{20wang,23universal}, the training dataset comprises 720k images (360k in real class, 360k in fake class), where the images in fake class are generated by ProGAN~\cite{18progan} trained on 20 different categories of LSUN~\cite{15lsun}. During training, Gaussian blur and JPEG compression are applied as data augmentations with a probability of 0.1. The AdamW~\cite{19adamw} optimizer is used with an initial learning rate of 0.0006. The load balancing coefficient is set to 0.01.\\

\noindent\textbf{Generative models evaluated.}
We evaluate the transferability of our method on 19 generators from the UnivFD dataset~\cite{23universal}, 8 generators from the GenImage dataset~\cite{23genimage}, and DALL-E 2~\cite{22dalle2} test set released by~\cite{23dire}. The UnivFD dataset includes the generators from~\cite{20wang}: ProGAN~\cite{18progan}, CycleGAN~\cite{17cyclegan}, BigGAN~\cite{19biggan}, StyleGAN~\cite{19stylegan}, GauGAN~\cite{19gaugan}, StarGAN~\cite{18stargan}, Deepfakes~\cite{19ffpp}, SITD~\cite{18sitd}, SAN~\cite{19san}, CRN~\cite{17crn}, IMLE~\cite{19imle}. In addition, it has been augmented with three diffusion models (Guided diffusion model (ADM)~\cite{21adm}, Latent diffusion model (LDM)~\cite{22sd}, GLIDE~\cite{22glide}) and one autoregressive model (DALL-E~\cite{21dalle}). We also evaluate the robustness of our method to post-processing operations on this dataset. The GenImage dataset primarily contains more recent diffusion-based generators: Midjourney~\cite{midjourney}, Stable Diffusion v1.4/v1.5~\cite{22sd}, ADM~\cite{21adm}, GLIDE~\cite{22glide}, Wukong~\cite{wukong}, VQDM~\cite{22vqdm}, BigGAN~\cite{19biggan}.\\

\noindent\textbf{Evaluation metrics.}
We evaluate the detectors using average precision and classification accuracy as metrics, following~\cite{20wang,20patch,23universal,23dire}. The classification threshold is tuned on the held-out validation set from the training data (\textit{i.e.} ProGAN data). For the UnivFD dataset, mean average precision (mAP) and average accuracy (avg.Acc) are reported following established methods~\cite{23universal}. For the GenImage dataset, we only report average accuracy as in~\cite{23genimage}.

\subsection{Generalization across unseen generators}
\label{sec:generalize}

\noindent\textbf{Comparisons on UnivFD dataset.}
We evaluate against several state-of-the-art methods on cross-generator generalization. Our baselines include: (\romannumeral1) \textbf{CNNSpot}~\cite{20wang} fine-tunes a ResNet-50~\cite{16resnet} (pre-trained on ImageNet~\cite{09imagenet}) on ProGAN/LSUN images with Gaussian blur and JPEG compression (probability of 0.1). (\romannumeral2) \textbf{PatchForensics}~\cite{20patch} trains a fully-convolutional patch-level classifier with limited receptive fields using Xception~\cite{17xception}. (\romannumeral3) \textbf{CoOccurrence}~\cite{19cooccurrence} uses co-occurrence matrices to train a real-vs-fake image classifier. (\romannumeral4) \textbf{FreqSpec}~\cite{19zhang} detects sampling artifacts on frequency spectra in GAN-generated images. (\romannumeral5) \textbf{DIRE}~\cite{23dire} identifies diffusion-generated images using the reconstruction error of the pre-trained ADM~\cite{21adm}. We train it on ProGAN data as others based on official codes. (\romannumeral6) \textbf{UnivFD}~\cite{23universal} employs the pre-trained features of CLIP~\cite{21clip} visual encoder for classification via nearest neighbor and linear probing. We use its superior variant for comparison, \textit{i.e.} linear probing classifier. For fair comparison, all baselines are trained on the same dataset as ours (\textit{i.e.} ProGAN data from~\cite{20wang}), except FreqSpec~\cite{19zhang} using CycleGAN data.

\begin{table*}[htbp]
\centering
\resizebox{\linewidth}{!}{
\setlength{\tabcolsep}{3mm}{
\begin{tabular}{lcccccccccc}
\toprule
Method& BigGAN & ADM   & GLIDE & Midjourney & SD v1.4 & SD v1.5 & VQDM  & Wukong & DALL-E 2 & avg. Acc \\ \midrule
CNNSpot~\cite{20wang} & 71.17  & 60.39 & 58.07 & 51.39 & 50.57   & 50.53   & 56.46 & 51.03  & 50.45& 55.56    \\
FreqDCT~\cite{20frank} & 81.97  & 63.42 & 54.13 & 45.87 & 38.79   & 39.21   & 77.80 & 40.30  & 34.70& 52.91    \\
GramNet~\cite{20texture} & 67.33  & 58.61 & 54.50 & 50.02 & 51.70   & 52.16   & 52.86 & 50.76  & 49.25& 54.13    \\
LNP~\cite{22byreal} & 77.75  & \underline{84.73} & 80.52 & 65.55 & 85.55   & 85.67   & 74.46 & \underline{82.06}  & \underline{88.75}& 80.56    \\
LGrad~\cite{23lgrad} & 85.62  & 67.15 & 66.11 & 65.35 & 63.02   & 63.67   & 72.99 & 59.55  & 65.45& 67.66    \\
DIRE*~\cite{23dire}& 70.12  & 75.78 & 71.75 & 58.01 & 49.74   & 49.83   & 53.68 & 54.46  & 66.48& 61.09    \\
UnivFD~\cite{23universal} & 95.08  & 66.87 & 62.46 & 56.13 & 63.66   & 63.49   & 85.31 & 70.93  & 50.75& 68.30    \\
\midrule
\rowcolor[HTML]{EFEFEF}
Ours (ViT-B/32) & 93.80 & 82.00 & 79.50 & \underline{73.65} & \textbf{87.95} & \underline{85.95} & \underline{87.70} & 80.10 & 71.35 & 82.44 \\
\rowcolor[HTML]{EFEFEF}
Ours (ViT-B/16) & \underline{96.20} & 76.00 & \underline{85.35} & 72.15 & 81.25 & 82.35 & 86.20 & 82.00 & 83.85 & \underline{82.82} \\
\rowcolor[HTML]{EFEFEF}
Ours (ViT-L/14) & \textbf{98.95} & \textbf{85.70} & \textbf{94.10} & \textbf{77.45} & \underline{86.90} & \textbf{87.20} & \textbf{91.95} & \textbf{86.60} & \textbf{93.15} & \textbf{89.11} \\ \bottomrule
\end{tabular}}}

\caption{\textbf{Generalization results on GenImage dataset~\cite{23genimage} and DALL-E 2~\cite{22dalle2} test set released by~\cite{23dire}.} The methods are trained on ProGAN data from~\cite{20wang} and evaluated using the classification accuracy (Acc) metric. The best result is highlighted in bold, and the second-best is underlined. * denotes our implementation based on official codes.}
\label{tab:general_genimage_acc}
\end{table*}

\begin{figure*}[htbp]
    \centering
    \begin{subfigure}[b]{.495\textwidth}
        \centering
        \includegraphics[width=\textwidth]{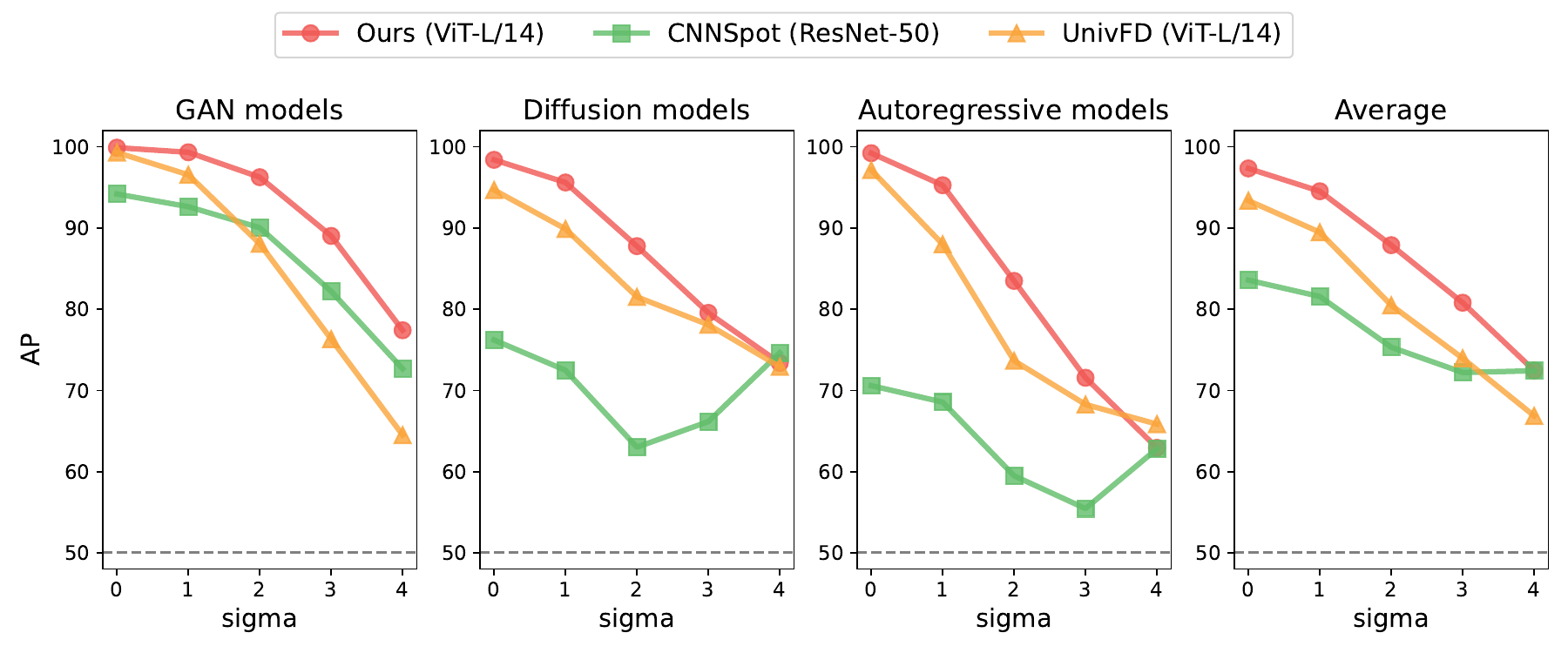}
        \caption{Robustness to Gaussian blur}
        \label{fig:robust_blur}
    \end{subfigure}
    \!
    \begin{subfigure}[b]{.495\textwidth}
        \centering
        \includegraphics[width=\textwidth]{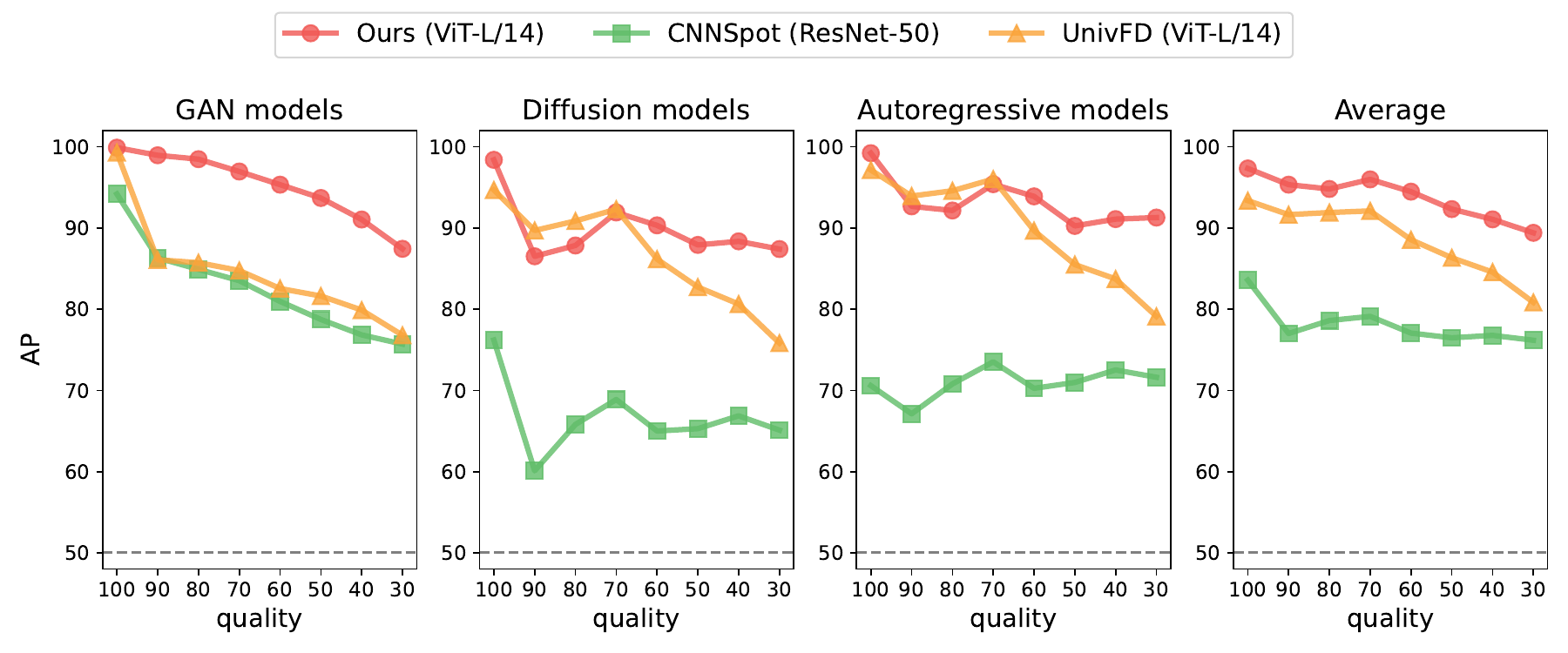}
        \caption{Robustness to JPEG compression}
        \label{fig:robust_jpeg}
    \end{subfigure}
    
    \caption{\textbf{Robustness to post-processing operations.} We assess the resilience of the detectors against two test-time perturbations, \textit{i.e.} \textbf{(a)} Gaussian blur and \textbf{(b)} JPEG compression, using the UnivFD dataset. We compare our method to CNNSpot and UnivFD, all trained under identical settings. We show the mAP results averaged across 19 generators and three types of generative models.}
    \label{fig:robust}
\end{figure*}

Table~\ref{tab:general_univfd_ap} and Table~\ref{tab:general_univfd_acc} present the average precision and classification accuracy results, respectively, of all detectors (rows) identifying fake images from various generators (columns). The shown accuracy values are averaged over an equal number of real and fake images for each generative model. We observe that while previous methods exhibit some generalization, their performances typically witness a drastic degradation when encountering unknown generators with substantial architectural discrepancies. For instance, CNNSpot~\cite{20wang} achieves 94.19\% mAP (85.29\% avg.Acc) on GANs but drops to 76.22\% mAP (59.06\% avg.Acc) on diffusion models. Recent works like DIRE~\cite{23dire} and UnivFD~\cite{23universal} offer promising solutions for handling unseen diffusion models. However, DIRE lacks consistent detection capability across diverse generators, with high performance on diffusion models (95.95\% mAP / 86.05\% avg.Acc) but lower on GANs (85.84\% mAP / 80.15\% avg.Acc), despite being trained on ProGAN data. In contrast, UnivFD consistently performs well, with 93.38\% mAP (81.38\% avg.Acc) across all settings, showing the efficacy of exploiting pre-trained vision-language models for this task. However, there remains room for improvement, as evidenced by its performance on low-level vision generators (70.17\% mAP / 60.25\% avg.Acc). Notably, our method significantly outperforms current state-of-the-art approaches. Specifically, our best-performing method (ViT-L/14) shows substantial improvements over UnivFD, achieving 97.45\% mAP / 92.45\% avg.Acc (+4.07\% mAP / +11.07\% avg.Acc) across all settings, and 98.64\% mAP / 94.74\% avg.Acc (+3.64\% mAP / +12.72\% avg.Acc) across unseen diffusion and autoregressive models (\textit{i.e.} LDM, GLIDE, Guided, and DALL-E). In conclusion, these results unequivocally highlight the advantage of our method, which incorporates mixture of low-rank experts to fully unleash the aptitude of CLIP-ViT.\\

\noindent\textbf{Comparisons on GenImage dataset.}
GenImage includes more recent advanced generators that are capable of producing super-realistic images, increasing the risk of their misuse for creating deceptive images that could seriously damage social media. Apart from some of the previously mentioned methods, we also evaluate several other strong baselines from this benchmark: (\romannumeral1) \textbf{FreqDCT}~\cite{20frank} uses a shallow CNN trained on DCT spectra to detect GAN-generated images. (\romannumeral2) \textbf{GramNet}~\cite{20texture} enhances fake face detection by integrating global texture extraction into ResNet. (\romannumeral3) \textbf{LNP}~\cite{22byreal} extracts learned noise patterns to detect fake images in frequency domain. (\romannumeral4) \textbf{LGrad}~\cite{23lgrad} obtains gradient maps as fingerprints using a pre-trained image classifier to filter out image content and perform classification. 

Table~\ref{tab:general_genimage_acc} reveals lower accuracies across all methods on the GenImage dataset, highlighting its increased challenge. Our method consistently outperforms the leading baseline, LNP~\cite{22byreal}, with the best variant (ViT-L/14) showing an 8.55\% avg.Acc improvement.

\subsection{Robustness to post-processing operations}
\label{sec:robust}
\begin{figure*}
    \centering
    \includegraphics[width=\textwidth]{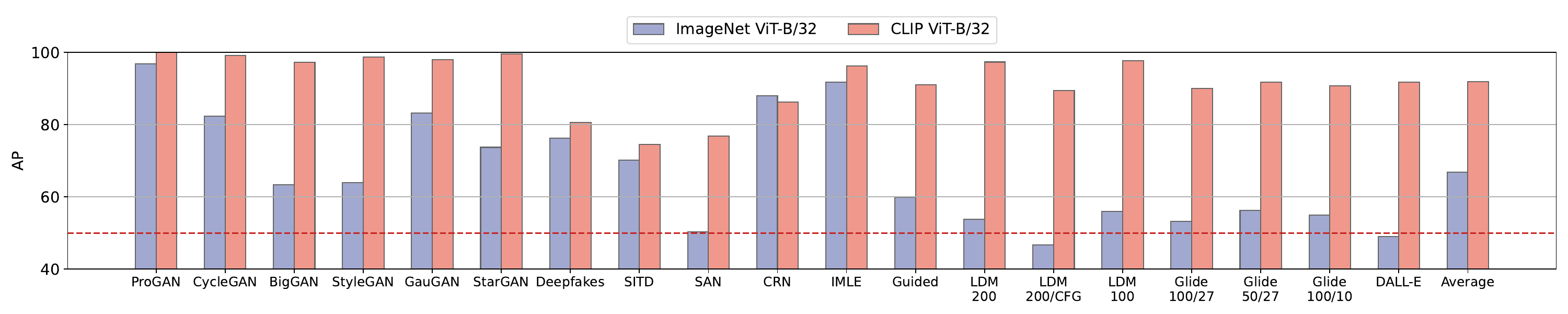}
    \caption{\textbf{Effect of pre-trained backbone.} The pre-trained ViT of CLIP substantially enhances generalization across nearly all generators when compared to that pre-trained on ImageNet classification. The red dotted line indicates chance performance.}
    \label{fig:backbone}
\end{figure*}

\begin{figure*}
    \centering
    \includegraphics[width=\textwidth]{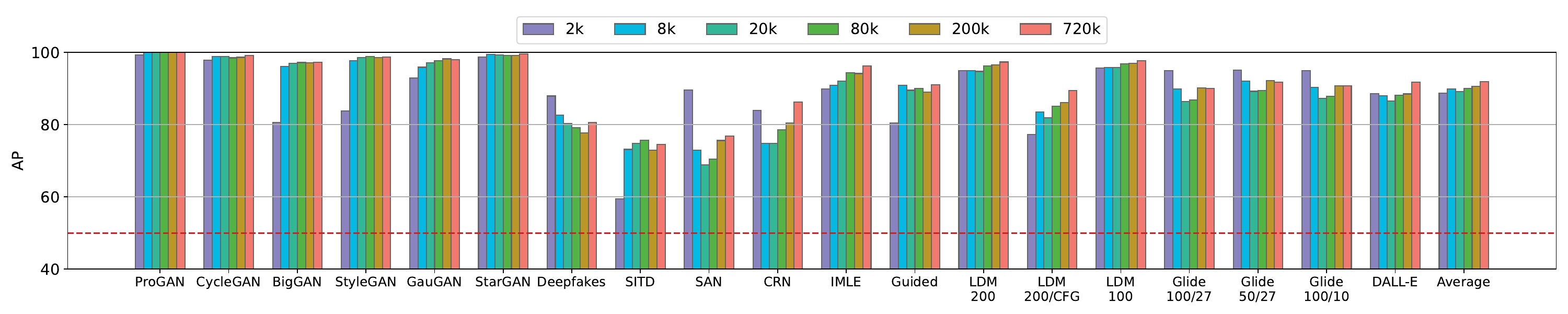}
    \caption{\textbf{Effect of training data size.} We evaluate the performance variation of our CLIP ViT-B/32 variant under different training data sizes, showing that our method can maintain strong generalization without requiring extensive training data.}
    \label{fig:data_size}
\end{figure*}

Real-world images often undergo various unknown degradations to evade detection systems. Hence, we assess the robustness of our approach on the UnivFD dataset perturbed by two common post-processing operations: Gaussian blur and JPEG compression, following~\cite{20wang,23universal,23dire}. The operations are applied at four levels for Gaussian blur (sigma = 1,2,3,4) and seven levels for JPEG compression (quality = 90,80,70,60,50,40,30). We compare the robustness of CNNSpot~\cite{20wang}, UnivFD~\cite{23universal}, and our method, all trained with Gaussian blur and JPEG compression augmentations on ProGAN data. Figure~\ref{fig:robust} shows the mAP results on the average of 19 generators and three types of generative models: GAN models (averaged over ProGAN, CycleGAN, \textit{etc.}), diffusion models (averaged over Guided, LDM, \textit{etc.}), and autoregressive model (DALL-E). We observe that all methods exhibit decreased performance when faced with blur and JPEG perturbations. This can be attributed to the elimination of low-level artifacts crucial for distinguishing fake from real images. CNNSpot shows more consistent performance across varied blur and JPEG levels, but its mAP is notably lower compared to UnivFD and ours. Our approach shows better overall robustness compared to UnivFD, as evidenced by the flatter robustness curve.

\subsection{Effect of pre-trained backbone}
We aim to enhance the generalization of AI-generated image detection using pre-trained visual knowledge from an open-world context. In this section, we compare the performance of ViTs under different pre-training setups. Consistent with~\cite{23universal}, we underscore the crucial role of the pre-trained backbone in generalization. We consider using ViT-B/32 pre-trained on ImageNet-21k~\cite{09imagenet} as a baseline while keeping other training settings the same. As shown in Figure~\ref{fig:backbone}, CLIP's pre-trained visual encoder significantly boosts generalization across nearly all generators (except CRN~\cite{17crn}), particularly on unseen diffusion models. The superiority of CLIP-ViT may stem from its exposure to a wider variety of images, thus better understanding the distribution of natural data. Additionally, the image-text contrastive pre-training task could strengthen the descriptive power of CLIP features and better model low-level image details necessary for distinguishing fake from real images.

\subsection{Effect of training data size}
We wonder whether our method can maintain such strong generalization without extensive training data, considering the minimal extra parameters introduced by our adapter modules. In this section, we evaluate the performance of our method under training data of different sizes, sampled uniformly from the 720k ProGAN/LSUN dataset (\{2k, 8k, 20k, 80k, 200k, 720k\}). Figure~\ref{fig:data_size} shows the performance variation of our ViT-B/32 variant across various generative models. Surprisingly, we observe only marginal performance degradation as the training data size decreases, indicating its resilience to data size variations and its ability to achieve strong generalization without large-scale training. Notably, the detector trained with just 2k data outperforms that trained on the full dataset in several generative models, \textit{i.e.} Deepfakes, SAN, and GLIDE. This suggests potential biases introduced by overtraining on domain-specific data, possibly hindering the detection of certain generative patterns and diminishing the advantages of CLIP features. Even with only 2k data ($\sim$0.28\% of the original dataset), our ViT-L/14 variant achieves an mAP of 94.25\% on the UnivFD dataset, surpassing the leading baseline UnivFD~\cite{23universal} (93.38\%).

\subsection{Ablation on adapted blocks}
\label{sec:ablation_block}

Do all ViT blocks need adaption? The key takeaway is that only deeper blocks require adaption to achieve optimal performance. Empirically, we adapt different sets of blocks to find the optimal configuration for ViT-B/32 (12 blocks) and ViT-L/14 (24 blocks), as in Table~\ref{tab:ablation_block}. Results show that adapting the last three blocks yields the best performance for both ViT-B/32 and ViT-L/14. Note that "None" actually refers to the UnivFD~\cite{23universal} method. We also provide the percentage of trainable parameters for each variant.

\begin{figure*}[h]
\section*{\centering{\Large{Mixture of Low-rank Experts for Transferable AI-Generated Image Detection}}}
\section*{\centering{\Large{\normalfont{Supplementary Material}}}}
\vspace{5mm}  

\subsection*{Expert feature visualization}
\label{sec:expert_vis}

\raggedright{
We adapt all blocks of CLIP ViT-B/32 and visualize the feature space of each LoRA expert at every block using t-SNE~\cite{08tsne}. Each feature vector is the summation of features extracted by both the frozen MLP and one of the LoRA experts. Figure~\ref{fig:expert_vis} shows the visualization of data points for real and fake images on the UnivFD dataset. We observe that experts in shallow blocks show considerable redundancy, while deeper experts learn different feature patterns.
}

\centering
\;\\\;\\\;\\
\begin{subfigure}[b]{\textwidth}
    \centering
    \includegraphics[width=\textwidth]{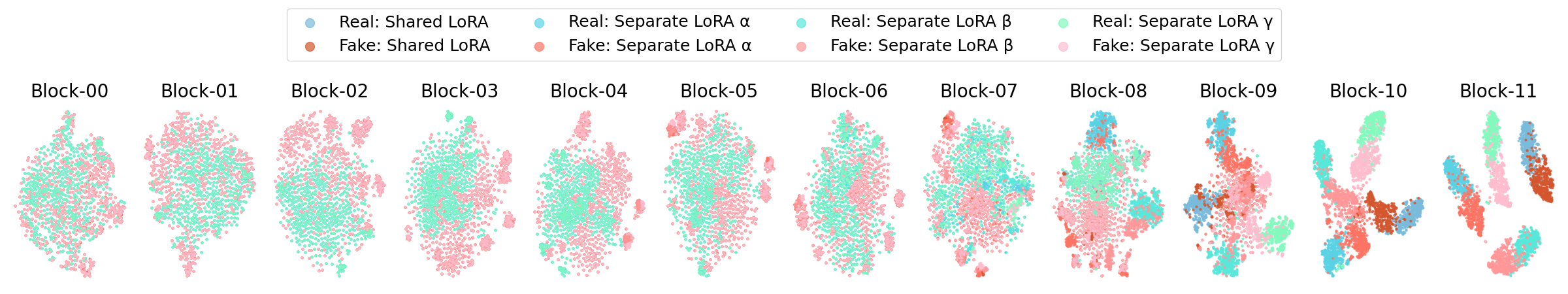}
    \caption{\textbf{Real vs Fake(All).} Fake images are uniformly sampled from all 19 generative models.}
    \label{fig:expertvis_real_fake}
\end{subfigure}
\\\;\\\;\\
\begin{subfigure}[b]{\textwidth}
    \centering
    \includegraphics[width=\textwidth]{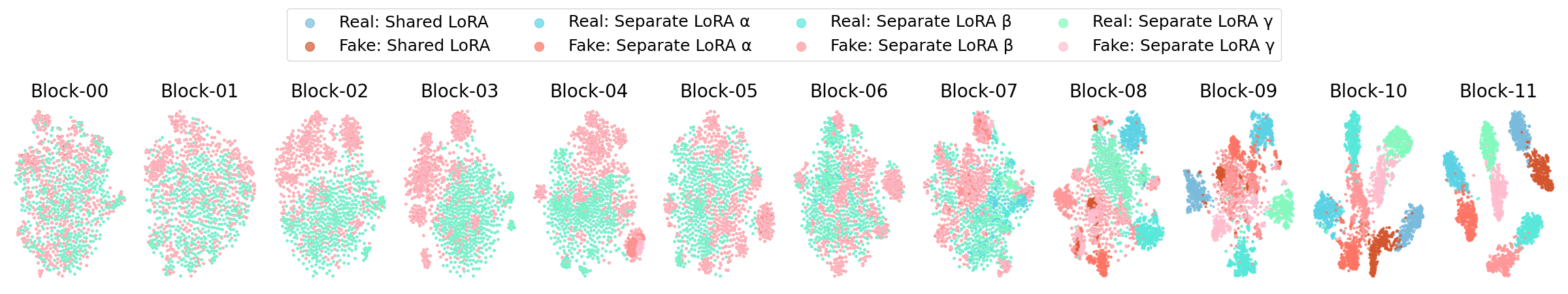}
    \caption{\textbf{Real vs Fake(GAN).} Fake images are uniformly sampled from all 6 GAN models.}
    \label{fig:expertvis_real_gan}
\end{subfigure}
\\\;\\\;\\
\begin{subfigure}[b]{\textwidth}
    \centering
    \includegraphics[width=\textwidth]{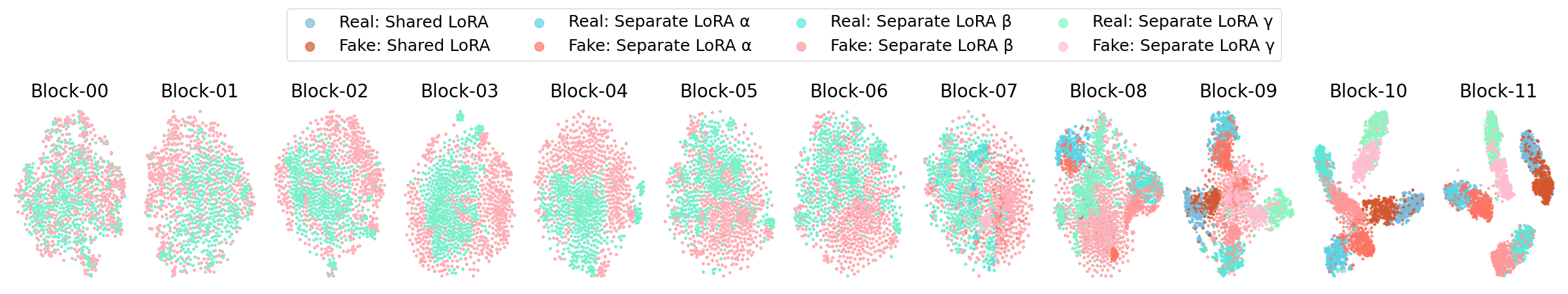}
    \caption{\textbf{Real vs Fake(Diffusion).} Fake images are uniformly sampled from all 7 diffusion models.}
    \label{fig:expertvis_real_dm}
\end{subfigure}
\\\;\\\;\\

\caption{t-SNE visualization of each LoRA expert at every block using the feature space of CLIP ViT-B/32 (with all blocks adapted). We visualize data points for real and fake images on the UnivFD dataset.}
\label{fig:expert_vis}
\end{figure*}

Moreover, we adapt all blocks of CLIP ViT-B/32 and visualize the feature space of each LoRA expert at every block using t-SNE~\cite{08tsne}, as in Figure~\ref{fig:expert_vis} (Supplementary Material). Shallow blocks in visual backbones typically capture low-level textural details, while deeper blocks focus more on the global semantics of images. Our goal is to leverage CLIP-ViT's strong low-level perception to extract forensic properties and feed them to deeper blocks for semantic analysis, facilitating the learning of a robust decision boundary. Optimizing shallow blocks on our data presents challenges in generalization for two main factors: (1) Shallow blocks may overfit to low-level artifacts from specific generative models in the training set to easily reduce the training loss, resulting in weak generalization as deeper blocks struggle to learn nontrivial patterns. (2) Figure~\ref{fig:expert_vis} shows that shallow block experts resemble each other, possibly due to representational collapse or learned routing policy overfitting~\cite{23moedropout}, making them more susceptible to overfitting. In contrast, deeper blocks require a more fine-grained manner to discern the semantic disparities between generated and natural images, a crucial aspect for strong generalization.

\subsection{Ablation on adapter module designs}
\label{sec:ablation_adapter}
\begin{table}[t]
\centering
\resizebox{\linewidth}{!}{
\begin{tabular}{ccccccc}
\toprule
\begin{tabular}[c]{@{}c@{}}CLIP\\ backbone\end{tabular} & \begin{tabular}[c]{@{}c@{}}Adapted\\ blocks\end{tabular} & \begin{tabular}[c]{@{}c@{}}\% Trainable\\ parameters\end{tabular} & \begin{tabular}[c]{@{}c@{}}GAN\\ models\end{tabular} & \begin{tabular}[c]{@{}c@{}}Diffusion\\ models\end{tabular} & \begin{tabular}[c]{@{}c@{}}Autoregressive\\ models\end{tabular} & Average \\ \midrule
\multirow{7}{*}{ViT-B/32} & None & 0.001\% & 93.69 & 88.15 & 89.65 & 87.62 \\ \cmidrule(l){2-7} 
 & Last 1 & 0.067\% & 97.49 & 90.92 & 91.74 & 90.81  \\ \cmidrule(l){2-7} 
 & Last 2 & 0.133\% & 98.78 & 92.58 & 92.76 & 91.57 \\ \cmidrule(l){2-7} 
 & Last 3 & 0.200\% & 99.02 & \textbf{94.79} & \textbf{96.64} & \textbf{92.39} \\ \cmidrule(l){2-7} 
 & Last 4 & 0.266\% & 99.20 & 93.08 & 93.94 & 91.95 \\ \cmidrule(l){2-7} 
 & Last 5 & 0.332\% & 99.26 & 91.71 & 92.18 & 91.19 \\ \cmidrule(l){2-7} 
 & All    & 0.792\% & \textbf{99.34} & 87.43 & 88.30 & 89.58 \\ \midrule
\multirow{4}{*}{ViT-L/14} & Last 2 & 0.051\% & 99.86 & 98.35 & 99.12 & 97.26 \\ \cmidrule(l){2-7} 
 & Last 3 & 0.077\% & \textbf{99.92} & \textbf{98.55} & \textbf{99.22} & \textbf{97.45}  \\ \cmidrule(l){2-7} 
 & Last 4 & 0.103\% & 99.81 & 98.15 & 98.76 & 97.36  \\ \cmidrule(l){2-7} 
 & Last 6 & 0.154\% & 99.85 & 96.80 & 97.92 & 96.42  \\ \bottomrule
\end{tabular}}
\caption{\textbf{Ablation on adapted blocks.} We present performance and the percentage of trainable parameters for each variant adapting different sets of blocks in ViT-B/32 and ViT-L/14.}
\label{tab:ablation_block}
\end{table}

\begin{table}[t]
\centering
\resizebox{\linewidth}{!}{
\begin{tabular}{clcccc}
\toprule
\multicolumn{2}{l}{Variant} & \begin{tabular}[c]{@{}c@{}}GAN\\ models\end{tabular} & \begin{tabular}[c]{@{}c@{}}Diffusion\\ models\end{tabular} & \begin{tabular}[c]{@{}c@{}}Autoregressive\\ models\end{tabular} & Average \\ \midrule
(a) & LoRA (MSA)                &  95.91 & 81.05 &  83.63 &  79.77 \\ \midrule
(b) & LoRA (MLP) / Shared LoRA  &  98.63 & 92.33 & 90.64 & 91.01 \\ \midrule
(c) & LoRA (MSA+MLP)            &  97.28 & 90.10 & 87.52 & 88.35 \\ \midrule
(d) & Mixture of separate LoRAs & 98.47 & 92.85 & 90.92 & 91.10 \\ \midrule
(e) & Ours                      & \textbf{99.02} & \textbf{94.79} & \textbf{96.64} & \textbf{92.39} \\ \bottomrule
\end{tabular}}
\caption{\textbf{Ablation on adapter module designs.} From (a), (b), (c), we explore fine-tuning MSA, MLP, and their combination with LoRA (rank=8). From (b), (d), (e), we investigate the effectiveness of shared and separate LoRA experts respectively. All experiments are conducted on CLIP ViT-B/32.}
\label{tab:ablation_adapter}
\end{table}

Why do we design our adapter modules in this manner? First, we conduct a preliminary study where we fine-tune CLIP ViT-B/32 with LoRA. We explore three variants, each fine-tuning different components with LoRA (rank=8) in the last three blocks: (a) the attention weights in multi-head self-attention (MSA) layers, \textit{i.e.} the query, key, value, and output projection matrices; (b) the projection matrices only in MLP layers, \textit{i.e.} shared LoRA only; (c) both MSA and MLP layers. Table~\ref{tab:ablation_adapter} shows that adapting only the MLP layers yields superior performance. Fine-tuning more parameters in self-attention leads to degraded performance instead. This could be attributed to MLP layers serving as key-value memories in Transformer-based models~\cite{21memory}, where CLIP-ViT stores its visual knowledge. However, self-attention layers learn cross-dependencies among patch tokens. Enhancing this ability appears unnecessary for our task, as patch-based classifiers already generalize well~\cite{20patch}.

\begin{figure*}[h]
\subsection*{Classification loss variation}
\label{sec:loss_curve}

\raggedright{
We visualize the variation of binary cross-entropy loss during training for the following variants: (b) Shared LoRA, (d) Mixture of separate LoRAs, and (e) Ours in Table~\ref{tab:ablation_adapter}. The experiments are conducted on a single GeForce RTX 2080 Ti GPU, with batch size set to 1024. We present the loss curves after applying moving averages with a window size of 10 (solid line) and before (transparent line). We observe that integrating both shared and separate adapters contributes to the model's convergence towards a globally better optimum.
}

\centering
\includegraphics[width=\textwidth]{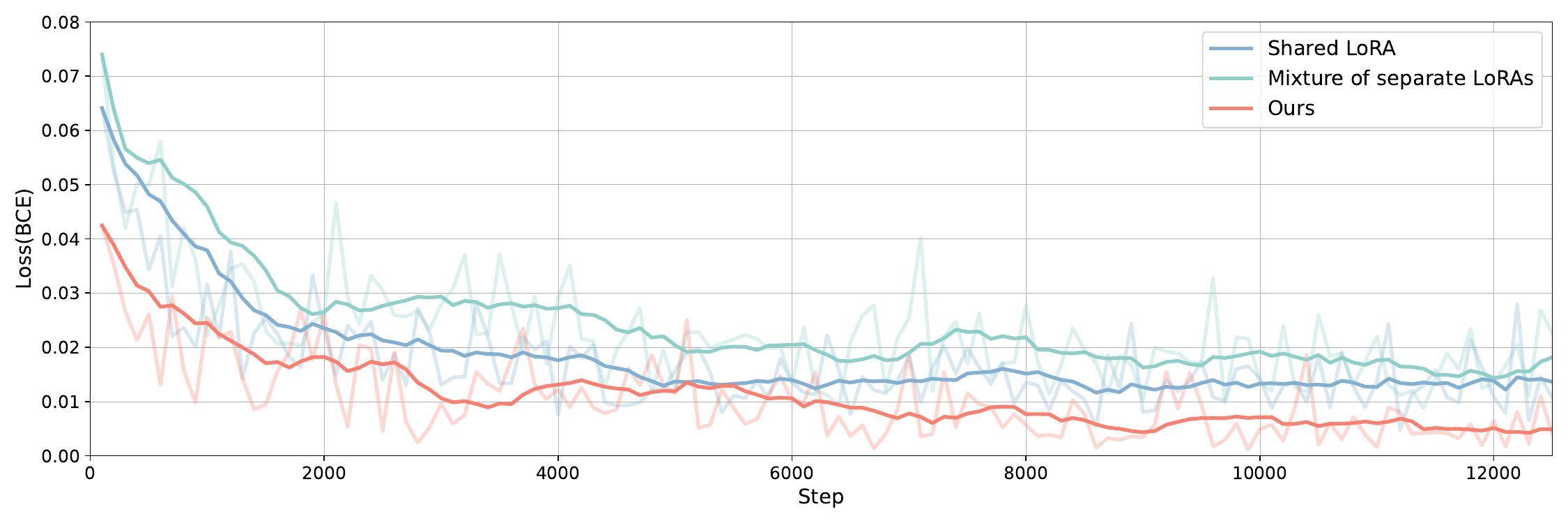}

\caption{The variation of binary cross-entropy loss during training for the variants: Shared LoRA, Mixture of separate LoRAs, and Ours. Solid lines denote the loss curves after moving averages of window size 10, while transparent lines represent the original curves.}

\label{fig:loss_curve}

\vspace{180mm}  

\end{figure*}

Further, we examine the efficacy of shared and separate adapters, and consider another variant (d) where MLP layers are adapted with only mixture of separate LoRAs (ranks=4,8,16). Results from Table~\ref{tab:ablation_adapter} show that shared and separate LoRA experts complement each other, leading to superior performance when combined compared to either used alone, as seen in rows (b), (d), and (e). Our conjecture is that the shared adapter learns common feature representations from the entire training set, such as intricate textures of natural images. Meanwhile, the sparse mixture of separate adapters encourages specialization in varied generative patterns of synthetic images, thereby enhancing the handling of diverse unknown generated images. We also visualize the variation of binary cross-entropy loss during training for the variants (b), (d), and (e) in Figure~\ref{fig:loss_curve} (Supplementary Material), which shows that integrating both shared and separate adapters contributes to the model converging towards a globally better optimum.

\subsection{Ablation on LoRA ranks}
\label{sec:ablation_rank}
\begin{table}[t]
\centering
\resizebox{\linewidth}{!}{
\begin{tabular}{c|cccc|cccc}
\toprule
\begin{tabular}[c]{@{}c@{}}Shared\\ LoRA rank\end{tabular} & \multicolumn{4}{c|}{\begin{tabular}[c]{@{}c@{}}Separate\\ LoRA ranks\end{tabular}} & \begin{tabular}[c]{@{}c@{}}GAN\\ models\end{tabular} & \begin{tabular}[c]{@{}c@{}}Diffusion\\ models\end{tabular} & \begin{tabular}[c]{@{}c@{}}Autoregressive\\ models\end{tabular} & Average \\ \midrule
4  & 4 & 8  & 16 &    & 98.30 & \textbf{94.86} & 93.19 & 92.22 \\ \midrule
8  & 4 & 8  & 16 &    & \textbf{99.02} & 94.79 & \textbf{96.64} & \textbf{92.39}  \\ \midrule
16 & 4 & 8  & 16 &    & 98.71 & 94.32 & 95.49 & 92.25 \\ \midrule
8  & 8 & 16 & 32 &    & 98.67 & 93.56 & 92.32 & 92.37 \\ \midrule
8  & 4 & 4  & 4  &    & 98.48 & 92.64 & 89.35 & 91.31 \\ \midrule
8  & 8 & 8  & 8  &    & 98.70 & 92.55 & 89.98 & 91.98 \\ \midrule
8  & 8 & 8  & 8  & 8  & 98.68 & 93.09 & 91.23 & 91.30 \\ \midrule
8  & 4 & 8  & 16 & 32 & 98.42 & 92.90 & 91.63 & 91.60 \\ \bottomrule
\end{tabular}}
\caption{\textbf{Ablation on LoRA ranks.} We examine the impact of various rank combinations of four LoRAs on CLIP ViT-B/32 and show consistently strong results.}
\label{tab:ablation_rank}
\end{table}

Do LoRA ranks significantly affect performance? Hyperparameter tuning experiments are performed to assess the impact of adjusting the ranks of the four LoRAs on CLIP ViT-B/32. Results in Table~\ref{tab:ablation_rank} show consistent performance across various rank combinations, indicating the stability of performance improvement with our proposed method.

\section{Conclusion}
\label{sec:conclusion}

This paper focuses on developing a transferable AI-generated image detector that enables universal detection across unknown sources. Current detectors exhibit significant room for improvement in terms of generalization and robustness, as demonstrated by several benchmarks. Drawing inspiration from the advantages of pre-trained CLIP, we aim to leverage its beneficial visual-world knowledge and descriptive proficiency to enhance out-of-domain generalization. We introduce a novel, parameter-efficient fine-tuning approach, mixture of low-rank experts, to fully exploit CLIP-ViT’s potential while preserving knowledge and expanding capacity for transferable detection. Extensive experiments indicate that our method outperforms state-of-the-art approaches in cross-generator generalization and robustness to perturbations. Our study highlights the presence of exploitable cues that can be consistently detected across a wide variety of generators, from early GANs to the latest diffusion models. However, the specific cues guiding the detector's decisions remain uncertain. We advocate for future research to explore interpretable forensics to design more effective detectors. We hope that our solution can not only establish strong baselines in this field, but also offer valuable insights for future works seeking to adapt vision-language models for broader visual recognition tasks.

{
    \small
    \bibliographystyle{ieeenat_fullname}
    \bibliography{main}
}

\clearpage
\setcounter{page}{1}

\end{document}